\documentclass[twocolumn,journal,10pt]{IEEEtran}
\usepackage{ifpdf, flushend,subfigure}
\DeclareUnicodeCharacter{3000}{ }
\DeclareUnicodeCharacter{3002}{ }

\def\bp{{\mathbf{p}}}
\def\bq{{\mathbf{q}}}
\def\bc{{\mathbf{c}}}

\ifCLASSINFOpdf
  \usepackage[pdftex]{graphicx}
  \graphicspath{{../pdf/}{../jpeg/}}
  \DeclareGraphicsExtensions{.pdf,.jpeg,.png}
\else
  \usepackage[dvips]{graphicx}
  \graphicspath{{../eps/}}
  \DeclareGraphicsExtensions{}
\fi
\usepackage{epstopdf}
\usepackage{multirow}
\usepackage{float}
\usepackage[cmex10]{amsmath}
\usepackage{amssymb,bm}
\usepackage{mathtools}
\usepackage{graphicx}
\usepackage{array}
\usepackage{mdwmath}
\usepackage{mdwtab}
\usepackage{eqparbox}
\usepackage{nomencl}
\usepackage{cite}
\usepackage{algorithm}
\usepackage{algorithmicx}
\usepackage[noend]{algpseudocode}

\algnewcommand{\Initialize}[1]{%
  \State \textbf{Initialize:}
  \Statex \hspace*{\algorithmicindent}\parbox[t]{.8\linewidth}{\raggedright #1}
}

\usepackage{makeidx}
\usepackage{ifthen}
\usepackage{subfigure}
\usepackage{caption}
\usepackage{xcolor}
\usepackage{amsthm}
\usepackage{amsfonts}
\usepackage{pdflscape}

\newcommand{\argmax}{\operatornamewithlimits{argmax}}
\newcommand{\beq}{\begin{equation}}
\newcommand{\eeq}{\end{equation}}
\newcommand{\beqn}{\begin{eqnarray}}
\newcommand{\eeqn}{\end{eqnarray}}
\newcommand{\beqno}{\begin{eqnarray*}}
\newcommand{\eeqno}{\end{eqnarray*}}
\newcommand{\bma}{\begin{displaymath}}
\newcommand{\ema}{\end{displaymath}}
\newcommand{\bnu}{\begin{enumerate}}
\newcommand{\enu}{\end{enumerate}}
\newcommand{\bce}{\begin{center}}
\newcommand{\ece}{\end{center}}
\newcommand{\btb}{\begin{tabular}}
\newcommand{\etb}{\end{tabular}}

\newtheorem{theorem}{Theorem}

\newtheorem{remark}{Remark}
\allowdisplaybreaks[2]

\begin{document}

\title{A Bandit Approach to Online Pricing for Heterogeneous Edge Resource Allocation}

\author{\IEEEauthorblockN{Jiaming Cheng$^*$,~\IEEEmembership{Student Member,~IEEE},
Duong Thuy Anh Nguyen$^*$,~\IEEEmembership{Student Member,~IEEE}, \\Lele Wang,~\IEEEmembership{Member,~IEEE},~Duong Tung Nguyen,~\IEEEmembership{Member,~IEEE}, Vijay K. Bhargava,~\IEEEmembership{Life~Fellow,~IEEE}}  
\thanks{Authors with $*$ have contributed equally to this work.} 
} 
\maketitle

\begin{abstract}
 Edge Computing (EC) offers a superior user experience by positioning cloud resources in close proximity to end users. 
 The challenge of allocating edge resources efficiently while maximizing profit for the EC platform remains a sophisticated problem, especially with the added complexity of the online arrival of resource requests.
 To address this challenge, we propose to cast the problem as a multi-armed bandit problem and develop two novel online pricing mechanisms, the Kullback-Leibler Upper Confidence Bound (KL-UCB) algorithm and the Min-Max Optimal algorithm, for heterogeneous edge resource allocation.
These mechanisms operate in real-time and do not require prior knowledge of demand distribution, which can be difficult to obtain in practice. 
 The proposed posted pricing schemes allow users to select and pay for their preferred resources, with the platform dynamically adjusting resource prices based on observed historical data.
 Numerical results show the advantages of the proposed mechanisms compared to several benchmark schemes derived from traditional bandit algorithms, including the Epsilon-Greedy, basic UCB, and Thompson Sampling algorithms.
\end{abstract}


\printnomenclature

\section{Introduction}


The proliferation of mobile devices and services has spurred the growth of new and innovative applications such as augmented reality (AR), virtual reality (VR), autonomous driving, real-time analytics, and the tactile Internet. 
Edge computing (EC)  has established itself as a crucial technology that works in conjunction with central clouds to fulfill the stringent requirements of new applications and services. 
By positioning computing resources at the network edge, in close proximity to end-users, devices, and sensors, EC effectively reduces bandwidth consumption while ensuring high reliability and supporting delay-sensitive applications \cite{duong_wfiot,Duong_TCC,tara22,ella2022}.


The efficient allocation of resources from geographically dispersed, heterogeneous edge nodes (ENs) with limited capacity to various users with diverse preferences is a crucial concern that must be addressed. 
In this paper, we examine the operation of an EC platform, which serves as the edge network infrastructure provider and manages the allocation of edge resources to meet the demands of customers, also referred to as resource buyers. 
The  buyers (e.g., service providers, AR/VR companies, vertical industries, and enterprises) can purchase resources from the platform to place their data and applications on various ENs to provide low-latency services to their users. 
While every buyer seeks powerful ENs that are located close to their users, the capacity of each EN is limited.

To effectively balance the demands of buyers with the constraints of the available resources at geographically dispersed edge nodes (ENs) and optimize the profit of the platform, we propose the use of dynamic pricing as a solution. In particular, we consider the setting where the platform offers multiple virtual machine (VM) instances, situated in diverse ENs, at varying prices to various buyers who arrive and make resource requests in real time. Our proposed platform differs from previous approaches in that it operates in a dynamic environment and does not necessitate prior knowledge of the user demand.
Furthermore, it takes into account that each buyer may have their own private valuation of the edge resources, which remains undisclosed to the platform.

\textbf{Related Work:} 
The utilization of VMs in cloud/edge computing  plays a pivotal role in fulfilling the required flexibility and scalability to cope with the ever-increasing demands of a mobile-centric environment. Amazon's Elastic Compute Cloud (EC2) \cite{EC2} and Microsoft Azure are examples of platforms that have emerged, offering customers an array of VM instances that can be tailored to their specific preferences and usage, enabling customers to dynamically optimize their computing resources. Despite the enhanced granularity of resource provisioning, static pricing policies are still commonly adopted, which lack market flexibility and efficiency, and pose a threat to the platform's profitability and customer satisfaction. 


Thus, creating an effective edge resources market involves addressing various challenges, with optimizing resource allocation and pricing mechanisms being the foremost priorities.
The optimization of EC pricing models has gained substantial research attention in recent years.
Reference \cite{double18}  examines double auction-based schemes, while \cite{Duong_TON} and \cite{Duong_TCC} explore market equilibrium approaches to ensure fair and efficient resource allocation to multiple budget-constrained services.
Additionally, \cite{jyan21} presents a pricing framework based on the Stackelberg game theory, and \cite{tara22} proposes a bilevel optimization model to solve the joint edge resource management and pricing problems. Nevertheless, these works typically assume a static environment and predetermined user demand.

In dynamic environments characterized by online user arrivals, several studies have explored online/dynamic pricing mechanisms in various contexts. For instance, \cite{jun20}  proposes an online mechanism for a crowdsensing system with uncertain task arrivals, while \cite{three21} employs a three-stage Stackelberg game approach to address time-varying scenarios with uncertainty and maximize long-term profit. 
Similarly, the dynamic pricing of cloud resources has been investigated, primarily following the introduction of Amazon EC2 spot instances \cite{EC2}. Various online pricing mechanisms have been developed, including online VM auction optimization \cite{zhang17}, online learning-based marketplace \cite{ofm19}, online combinatorial auction \cite{Danny20}, and online auction based on the price function \cite{zhang21}. However, these mechanisms rely on prior knowledge of user demands. 

Multi-armed bandit (MAB), an effective online learning and optimization framework with partial feedback, has recently been applied to pricing and resource allocation. Several bandit-based mechanisms have been proposed for general online pricing problems, such as the auction-based combinatorial MAB mechanism in \cite{auction21} and the bandit-based mechanism for identical items with time-sensitive valuations introduced in \cite{time22}.
For cloud/edge computing,  \cite{Chuan18} designs a bandit-based algorithm for cloud resource pricing that allows the purchase of one or multiple instances of a single product. Additionally,  \cite{Ghoorchian2021} employs a MAB model to assign each computational task to a single edge server. 
It is noteworthy that previous studies have solely focused on either distinct VM types in the same location or identical VM types in different geographic locations. Moreover, they have confined buyers to procuring either a single or several instances of the same product.

\textbf{Contributions:} 
Designing an efficient, \textit{prior-independent} online pricing mechanism poses significant challenges in balancing the exploration-exploitation trade-off.
In response, we propose a new MAB framework for online edge resource pricing. 
Unlike previous bandit-based online pricing mechanisms in cloud/edge computing, our approach takes into account the interdependence between the computing power of the virtual machines (VMs) and their geographic locations. This is particularly important for delay-sensitive applications, as the buyers' valuations towards the edge resources are influenced by both the VM type and the location of the VM.
 Furthermore, our model allows buyers to purchase multiple products, unlike existing works that restrict buyers to one product.

We develop two \textit{distribution-free} algorithms, including the Kullback-Leibler Upper Confidence Bound (UCB) \cite{KLUCB}  and the  min-max optimal strategy (MOSS)  \cite{MOSS}, for online edge resource pricing. 
These algorithms operate in real-time and eliminate the need for prior knowledge of the demand distribution and ensure truthfulness as the online posted price is independent of the newly arrived buyer's valuation. 
We evaluate the performance of our proposed mechanisms by comparing them with 
traditional bandit algorithms such as Epsilon-Greedy \cite{Epislongreedy}, basic UCB \cite{UCB}, and Thompson Sampling \cite{TS}. The performance is measured using the concept of regret, which is the difference between the expected reward of the best arm and the expected reward of the selected arms using our pricing strategies. Our goal is to minimize regret and optimize revenue and resource utilization.

The remaining paper is organized as follows. Section  \ref{model} presents the system model and problem formulation. Section \ref{sec:sol} provides the solution approach. Finally, Section \ref{simulation} shows the simulation results, followed by conclusions  in Section \ref{summary}.

\section{System Model and Problem formulation}
\label{model}

\subsection{Edge Resource Allocation and Pricing Problem}
We consider an edge resource allocation problem for an EC system with a platform that 
manages a set of $N$ resource-constrained heterogeneous ENs, denoted as $\mathcal{N}=\{1,\ldots,N\}$, located in different areas, to provide computing resources in the form of virtual machines (VMs). In each EN, there is a set $\mathcal{M}=\{1,\ldots, M\}$ of $M$ types of VM instances available to serve users, each with different resource configurations in terms of  vCPU, memory, and storage \cite{EC2}. 
The platform offers a total of $MN$ different products, each represented by a tuple $(i,j)$ with $i \in \mathcal{M}$ and $j \in \mathcal{N}$. Each tuple represents VM type $i$ on EN $j$. By considering both VM types and their physical locations, this approach is particularly useful for  buyers with delay-sensitive applications. 

We consider  a set $\mathcal{T}=\{1,\ldots,T\}$ of $T$ buyers that arrive and request edge resources in an online manner.
Each buyer $t$ has unique computing tasks, which may require VMs in specific locations to meet latency requirements.
This leads to varying valuations for different products $(i,j)$, which are captured by the valuation function $v_{i,j}^{t}$. 
The \textit{valuation function for each buyer is private and unknown to the platform}.
When a buyer $t$ arrives, the EC platform updates the price $p_{i,j}^t$ for each product $(i,j)$, and the buyer determines their demand based on the prices set by the platform. While a buyer can purchase multiple products, they are permitted to buy at most one unit of each product. Each buyer aims to maximize her utility, while 
the platform's objective is to maximize the total revenue generated from selling resources to buyers.

\subsection{Dynamic Edge Resource Pricing as a MAB Problem}
Our goal is to design online mechanisms that allow the platform to make online decisions with performance guarantees. Since no prior information about the valuation functions of buyers is available to the platform, we approach this problem by casting it as a MAB problem. Specifically, the platform 
selects a single price vector of the products to offer to a newly arrived buyer at each time $t$, resembling the act of pulling a single arm from a set of available arms (prices). The outcome, which can either be a purchase or a refusal to buy, is then observed and accompanied by a reward.

In MAB problems, the decision maker is limited to a finite set of choices, known as the set of arms. However, in an online pricing problem, the action space for pricing may be very large or even infinite. To overcome this, we take advantage of the structured nature of the action space, where prices are simply numbers within a fixed interval, and discretize the action space by assuming the existence of a predefined set of discrete product prices at each EN, i.e., we have:
$$p^t_{i,j} \in \{p_{i,j}^{t,1} , \ldots, p_{i,j}^{t,V} \}, ~ \forall i,j,t,$$
where  $v \in \{1, \ldots, V\}$ represents different price options $p_{i,j}^{t,1} < p_{i,j}^{t,2} < \ldots < p_{i,j}^{t,V}$. The assumption is reasonable given that the price options can represent a variety of price levels (e.g., very low price, low price, medium price, high price, very high price). This approach is referred to as pre-adjusted discretization, a concept that has been explored in various literature. In our formulation, we consider a fixed and known set of $K$ arms, each representing a possible price vector, denoted by $\mathcal{P}$ with a cardinality of $|\mathcal{P}|=K$.

The algorithm is the monopolist EC platform that interacts with a set $\mathcal{T}$ of $T$ potential buyers whose requests are arriving one by one. The time proceeds in $T$ rounds as buyers arrive, where $T$ is a finite, known time horizon. In each round $t$, buyer $t$ arrives, the algorithm picks an arm $\bp^t\in \mathcal{P}$, i.e., a vector of prices $\bp^t=(p_{1,1}^{t},\ldots,p_{1,N}^{t},p_{2,1}^{t},\ldots,p_{M,N}^{t})$ 
and offers at most one unit of each VM $(i,j)$ at price $p_{i,j}^{t}$ to buyer $t$. Recalling that each buyer $t$ has its own valuations $v_{i,j}^{t}$ to each of the offered product type $(i,j)$. The buyer then chooses to procure a subset of products based on their valuations and leaves. In particular, the buyer purchases the product $(i,j)$, i.e, VM type $i$ at EN $j$, if $v_{i,j}^{t} \geq p_{i,j}^{t}$. The valuation function is assumed to be drawn \textit{from a fixed (but unknown) distribution} over the possible valuation functions, called the demand distribution.  

Once the buyer $t$ decides on the set of products to buy at the offered prices, the platform receives the payment from the buyer, i.e., the reward $r^t$ which is assumed to be bounded in $[0,1]$ ($r^t \in [a,b]$ can be scaled to satisfy the assumed bound). The EC platform then allocates the requested resources to the buyer $t$, and consumes the amount $c_{i,j}^t \in \{0,1\}$ of product $(i,j)$. This implies:
\begin{equation}
\begin{cases}
  c_{i,j}^t =1~~ \text{if} ~~ v_{i,j}^{t} \geq  p_{i,j}^{t},\\
  c_{i,j}^t = 0 ~~ \text{if} ~~ 0 \leq v_{i,j}^{t} < p_{i,j}^{t}.
\end{cases}
\end{equation}
Then, the utility of buyer $t$ can be expressed as $u_{t} = \sum_{i=1}^{M}\sum_{j=1}^{N} (v_{i,j}^{t} - p_{i,j}^{t}) c_{i,j}^{t}$. For clarity, we denote the following consumption vector for the resources including $MN$ products at round $t$:
\[ \bc^t=(c_{1,1}^{t},\ldots,c_{1,N}^{t},c_{2,1}^{t},\ldots,c_{M,N}^{t}) \in \{0,1\}^{MN}.\]
We define outcome vector $(r^t; \bc^t) \in [0,1]\times \{0,1\}^{MN}$ for round $t$ that includes the reward $r^t$ and resource consumption vector $\bc^t$. The values of $r^t$ and $\bc^t$ can be only revealed to the platform after the buyer makes decisions at round $t$.

The platform aims to determine a ``\textit{policy}" to associate its next decisions $\bp^{t}$ using past observations $\left((\bp^{1},r^1), (\bp^{2},r^2), \dots, (\bp^{t-1},r^{t-1})\right)$. The optimal decisions is the set of price vector $\left\{\bp^{t*}:t = 1,2,\dots\right\}$ with maximal expected reward $\mu^{*}$ . The performance of the policy is evaluated using the \textit{regret}, which measures the difference between the expected reward of the best arm and the expected reward of the selected arms. We denote $\textit{REG}_{t}$ as the regret of the policy with respect to the optimal policy evaluated at time $t$.

\section{Solution Approaches} \label{sec:sol}
By leveraging the classical UCB algorithm \cite{UCB},
we develop two online posted-price mechanisms, namely Kullback-Leibler UCB (KL-UCB) and min-max optimal strategy (MOSS). KL-UCB is a model-based technique that focuses on upper-bounding the expected reward, while MOSS is geared towards minimizing the regret. Specifically, the KL-UCB algorithm calculates an upper confidence bound on the expected reward for each arm and selects the arm with the highest bound in each round. MOSS, on the other hand, determines the minimum regret for each arm by considering the worst-case scenario and selects the arm with the minimum regret in each round.

\subsection{KL-UCB Algorithm}
We present the KL-UCB algorithm, a variant of the popular UCB algorithm \cite{UCB}, in the context of  dynamic edge resource pricing. We analyze the regret of KL-UCB, which uses the Kullback-Leibler (KL) divergence as a measure of uncertainty and reaches the lower bound of Lai and Robbins \cite{laibound} in the special case of Bernoulli rewards. For arbitrary bounded rewards, KL-UCB is the only method that satisfies a uniformly better regret bound than the basic UCB policy. First, we define the following empirical reward of price vector $\bp$ at time $t$: 
\[\hat{r}^t(\bp) = \frac{R_t}{n_{t}(\bp)} =  \frac{\sum_{\tau = 1}^{t} r^\tau(\bp) }{n_{t}(\bp)},\] 
where $n_{t}(\bp)$ denotes the number of times the price vector $\bp$ has been played up to time $t$ and $R_{t}=\sum_{\tau = 1}^{t} r^\tau(\bp)$ denotes the accumulative sum of reward up to time $t$.

Departure from UCB in \cite{UCB}, KL-UCB utilizes a distinct form of UCB estimates, which results in different regret bounds. The algorithm selects the available price vector $\bp$ with the highest $\textbf{UCB}^{\text{KL}}_{\bp,t}$, which is defined as:
\begin{align*}
\label{updates}
    \textbf{UCB}^{\text{KL}}_{\bp,t} = \max \left\{\bq \in [0,1]: d \left(\hat{r}^{t}(\bp), \bq \right) n_t(\bp) \leq f(t) \right\},
\end{align*}
where the level of confidence is set by the exploration function $f(t) = \log(t) + \gamma \log(\log(t))$. Here, $d(\cdot,\cdot)$ is the KL divergence between two probability distributions, the estimated reward distribution and the prior distribution. If the inputs are vectors, it is understood as being computed component-wise. For example, KL divergence between Bernoulli distributions of parameters $u$ and $v$ is represented by $d(u,v) = u \log \frac{u}{v} + (1 - u) \log \frac{1 - u}{1 - v}$. It is important to note that $d(u,v)$ is strictly convex and increasing on the interval $[u,1]$ for any $u \in [0,1]$. The hyper-parameter $\gamma$ is chosen to be equal to $0$ in practice for optimal results. Algorithm \ref{Alg:KL-UCB} provides the pseudocode for the KL-UCB algorithm.

\begin{algorithm}[h!]
\caption{Kullback-Leibler UCB (KL-UCB)}\label{Alg:KL-UCB}
\begin{algorithmic}[1]
    \State Try each arm once
    \For{$t = K+1$ to $T$ (i.e., until resource exhausted)}
        \State pick a price vector $\bp^t = \argmax_{\bp \in \mathcal{P}} \textbf{UCB}^{\text{KL}}_{\bp,t}$ 
        \State observe the consumption $\bc^t$ and reward $r^t$. 
    \EndFor
\end{algorithmic}
\end{algorithm}

Let $d({\mu_{\bp}},{\mu_{\bp^*}})$ be the KL divergence between estimated reward $\mu_{\bp}$ from choosing arm $\bp$ and the maximal expected reward $\mu_{\bp^*}$ from choosing the optimal arm $\bp^*$. For any integer $t>0$, we have the following bounds for the number of draws up to time $t$, $n_t(\bp)$, of any sub-optimal arm $\bp$ (see \cite{KLUCB}): 

\begin{theorem}
\label{Regret:lower-bound}
Assume that the distribution of observed reward for the arm $\bp$ belongs to a family $\{p_{\theta,\theta \in \Theta_{\bp}}\}$ of distributions. For any uniformly good policy, $n_t(\bp)$ is lower bounded by 
\begin{align} 
n_t(\bp) \!\!\geq \!\! \bigg( \frac{1}{\inf_{\theta \in \Theta_{\bp}: \mathbb{E}[p_{\theta_\bp}] > \mu_{\bp^{*}}} d(\mu_{\bp}, \mu_{\bp^{*}})} \!+\! o(1)\bigg)\! \log(k),
\end{align}
where $\mathbb{E}(p_\theta)$ is the expectation under $p_\theta$; hence, the regret is lower bounded as follows
\begin{align} 
\!\!\liminf\limits_{t \rightarrow \infty} \frac{\mathbb{E} [\textit{REG}_{t}]}{\log(t)} \!\geq \!\!\sum_{\bp: \mu_{\bp} \leq \mu_{\bp^{*}}}  \!\!\frac{\mu_{\bp^{*}} - \mu_{\bp}}{\inf_{\theta \in \Theta_{\bp}: \mathbb{E}[p_{\theta_\bp}] > \mu_{\bp^{*}}}\!\!d(\mu_{\bp}, \mu_{\bp^{*}})}.\!\!
\end{align}
\end{theorem}
\begin{remark}[Corollary 3 of \cite{KLUCB}]
The KL-UCB pricing scheme is asymptotically optimal if $r^{t}$ follows Bernoulli distribution \cite{laibound}: 
\[n_t(\bp) \geq \left(\frac{1}{d(\mu_{\bp}, \mu_{\bp^{*}})} + o(1)\right) \log(t),\] 
with a probability tending to $1$.
\end{remark}

\begin{theorem}[Theorem 2 of \cite{KLUCB}]
\label{Regret:upper-bound}
For the KL-UCB algorithm, let $\gamma = 3$ in Algorithm~\ref{Alg:KL-UCB},
$n_t(\bp)$ is upper-bounded by 
\begin{align} 
\mathbb{E}[n_{t}(\bp)] \!\leq \!\frac{\log(t)}{d(\mu_{\bp},\mu_{\bp^{*}})}(1 + \epsilon) \!+\! C_1 \log(\log(t)) \!+\! \frac{C_2(\epsilon)}{t^{b(\epsilon)}}
\end{align}
where $C_1$ is a positive constant and $C_2(\epsilon)$ and $b(\epsilon)$ denote positive function of $\epsilon>0$. Hence, 
\begin{align} 
\limsup\limits_{t \rightarrow \infty} \frac{\mathbb{E} [\textit{REG}_{t}]}{\log(t)} \leq \sum_{\bp: \mu_{\bp} \leq \mu_{\bp^{*}}} \frac{\mu_{\bp^{*}} - \mu_{\bp}}{d(\mu_{\bp}, \mu_{\bp^{*})}}.
\end{align}
\end{theorem}

Compared to the UCB algorithm \cite{UCB}, KL-UCB has a strictly better theoretical guarantee as the divergence $d({\mu_{\bp}},{\mu_{\bp^*}})>2(\mu_{\bp}-\mu_{\bp^*})^2$. This superiority has also been observed in simulations. KL-UCB can be easily adapted to handle other reward distributions by choosing a proper divergence function $d(\cdot,\cdot)$. For exponential rewards, the choice of divergence function should be $d(u,v)=\frac{v}{u}-1-\log(\frac{u}{v})$.

\subsection{Min-max optimal strategy (MOSS)}
In this section, we present the MOSS algorithm, which has been proven to achieve the best distribution-free regret of $\sqrt{TK}$ for stochastic bandits \cite{MOSS}. MOSS utilizes a unique UCB based on the empirical mean reward, defined as follows:
\begin{align} 
      \text{UCB}^{\text{MOSS}}_{\bp,t} = \hat{r}^t(\bp)  + \sqrt{\frac{\max \bigg( \log (\frac{T}{K  n_{t}(\bp)}),0 \bigg) }{n_{t}(\bp)}},
\end{align}
where $\hat{r}^t(\bp)$ refers to the empirical reward of arm $\bp$ at time $t$ and $n_{t}(\bp)$ represents the number of times arm $\bp$ has been played. The platform chooses the price vector that maximizes $\text{UCB}^{\text{MOSS}}_{\bp,t}$ at each time step. Algorithm \ref{Alg:MOSS} provides the pseudocode for the MOSS algorithm.

\begin{algorithm}[h!]
\label{ALG:MOSS}
\caption{Min-max optimal strategy in the stochastic case (MOSS) }\label{Alg:MOSS}
\begin{algorithmic}[1]
    \State Try each arm once
    \For{$t = K+1$ to $T$ (i.e., until resource exhausted)}
        \State pick a price vector $\bp^t = \argmax_{\bp \in \mathcal{P}} \text{UCB}^{\text{MOSS}}_{\bp,k}$ 
        \State observe the consumption $\bc^t$ and reward $r^t$. 
    \EndFor
\end{algorithmic}
\end{algorithm}

The following result states the bound for the regret of the distribution-free MOSS policy.
\begin{theorem}[Theorem 5 of \cite{MOSS}] 
\label{Regret:moss}
MOSS satisfies 
\begin{align}
\sup \textit{REG}_{T} \leq 49 \sqrt{TK},
\end{align}
where the supremum is taken over all $K$-tuple of probability distributions on $[0,1]$.
\end{theorem}

\section{Simulation}
\label{simulation}

\begin{figure}[ht!]
		 \subfigure[Average cumulative reward]{
	     \includegraphics[width=0.245\textwidth,height=0.12\textheight]{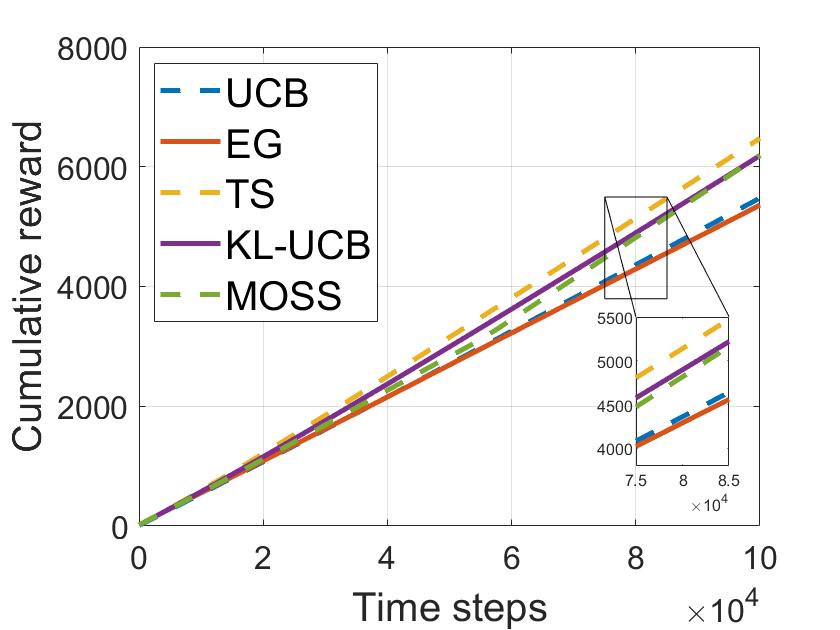}
	     \label{fig:Uniform_reward}
	}   \hspace*{-2.1em} 
	    \subfigure[Average cumulative regret]{
	     \includegraphics[width=0.245\textwidth,height=0.12\textheight]{ 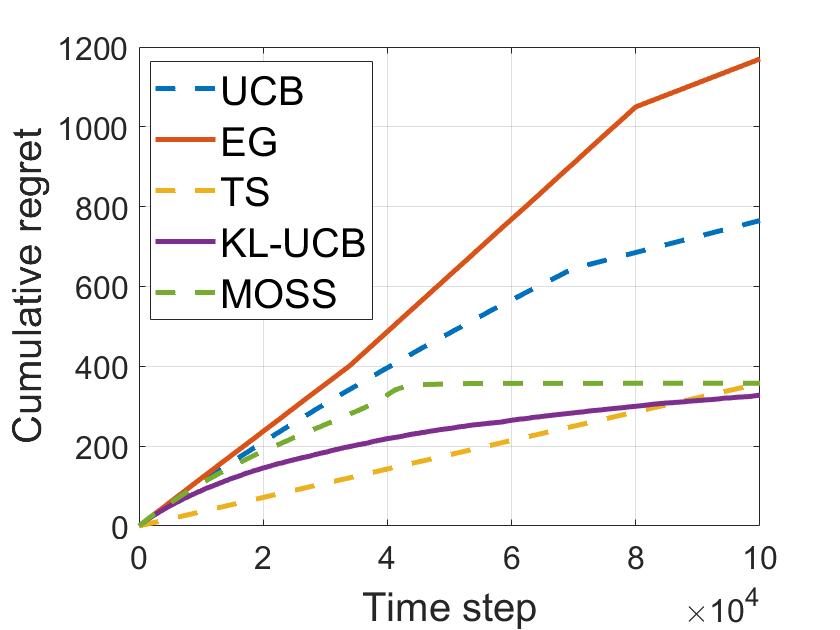}
	     \label{fig:Uniform_regret}
	} 
	    \caption{Uniform valuations}
     \label{fig:Uniform}
     \vspace{-0.3cm}
\end{figure}
\begin{figure}[ht!]
		 \subfigure[Average cumulative reward]{
	     \includegraphics[width=0.245\textwidth,height=0.12\textheight]{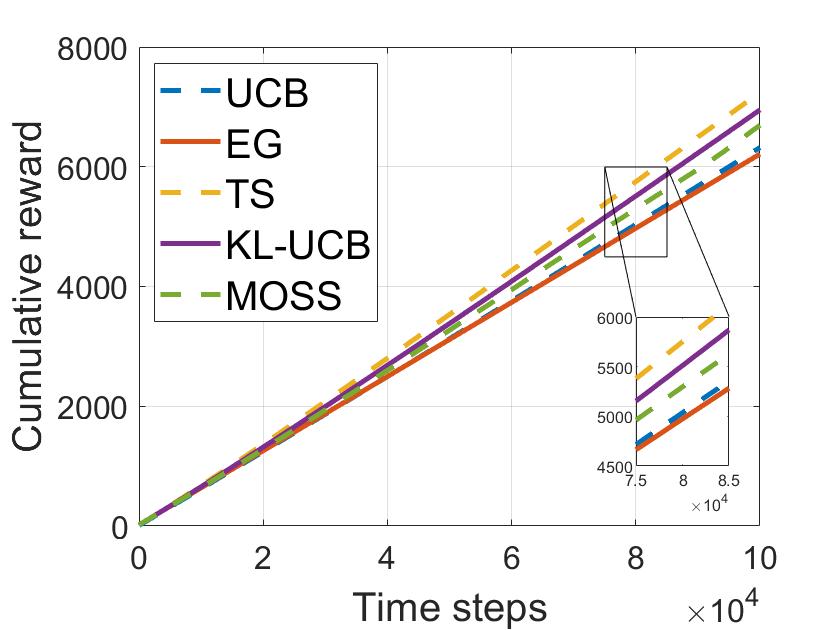}
	     \label{fig:Gaussian_avg_reward}
	}   \hspace*{-2.1em} 
	    \subfigure[Average cumulative regret]{
	     \includegraphics[width=0.245\textwidth,height=0.12\textheight]{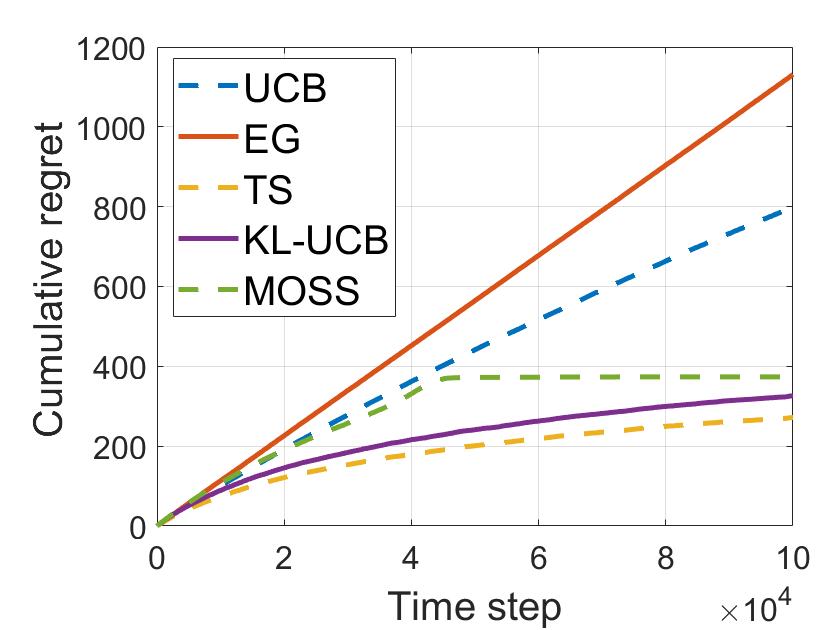}
	     \label{fig:Guassian_regret}
	} 
	    \caption{Gaussian valuations }
     \label{fig:Guassian}
     \vspace{-0.3cm}
\end{figure}
\begin{figure}[ht!]
		 \subfigure[Average cumulative reward]{
	     \includegraphics[width=0.245\textwidth,height=0.12\textheight]{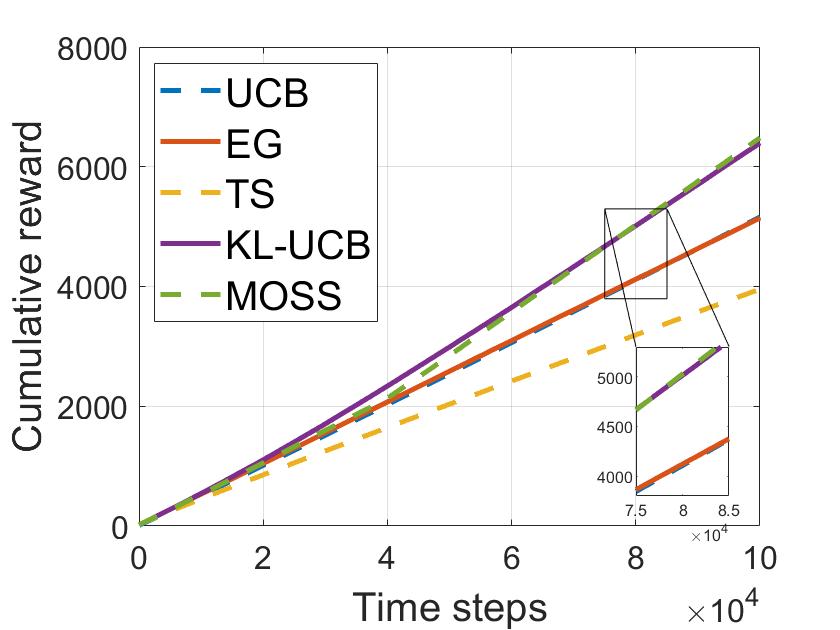}
	     \label{fig:exp_avg_reward}
	}   \hspace*{-2.1em} 
	    \subfigure[Arm selections]{
	     \includegraphics[width=0.245\textwidth,height=0.12\textheight]{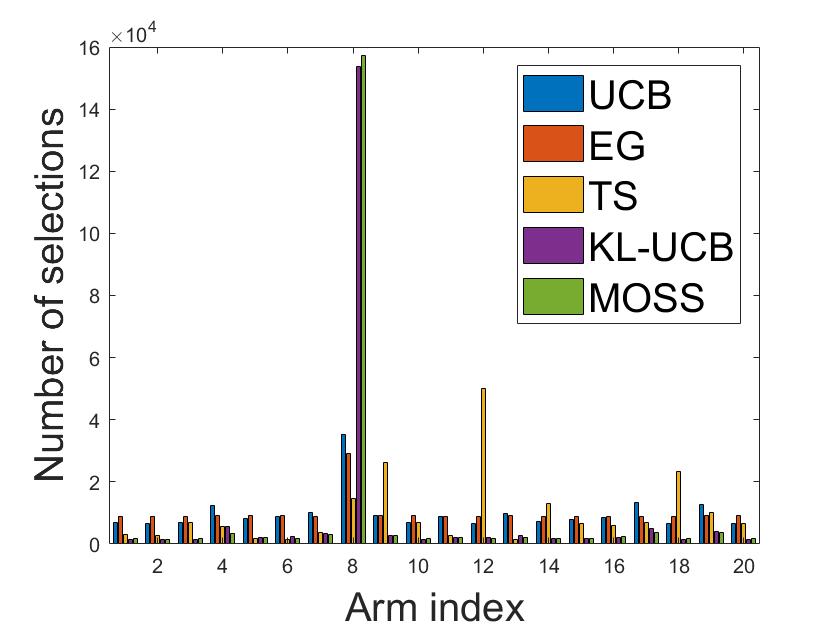}
	     \label{fig:exp_arms}
	} 
	    \caption{Exponential valuations}
     \label{fig:exp}
     \vspace{-0.3cm}
\end{figure}

In this section, we  evaluate the effectiveness  of the proposed mechanisms by comparing them with  traditional bandit algorithms, namely the Epsilon-Greedy \cite{Epislongreedy}, UCB \cite{UCB}, and Thompson Sampling (TS) algorithms. We present these algorithms in our technical report \cite{report}. 
Note that the proposed mechanisms are distribution-free and do not require prior knowledge of the buyers' valuations. Thus, to assess the performance of the other algorithms, we simulate buyer valuations as they make decisions based on their preferences for heterogeneous edge resources. These preferences may be influenced by factors such as network delay and computational requirements.  
 The platform updates the resource prices by observing and learning from the historical data and offers a take-it-or-leave-it price to each new buyer that arrives in an online fashion. Then, the newly arrived buyer chooses the product that offers the highest value for her money to maximize her utility.
In our simulation, we consider  $N = 3$ ENs and $M =3$ types of VM. The platform interacts with $T=100000$ buyers and has $K = 20$ pricing options \cite{EC2} to choose from in each interaction. 
We model the environment in which each buyer's valuation ($v_{t}$) is independently and identically generated from a truncated distribution. We consider the following distributions: 
\begin{itemize}
    \item[(1)] \textbf{Uniform}: $U[0,1]$. 
    \item[(2)] \textbf{Gaussian:} with mean $\mu = 0.2$ and variance $\sigma = 0.2$.
    \item[(3)] \textbf{Exponential:} with mean $\mu = \frac{1}{\lambda} = 2$.
\end{itemize}
We run $1000$ episodes for each distribution setting and compute the cumulative reward and cumulative regret.
The results 
are depicted in Figure~\ref{fig:Uniform}, Figure~\ref{fig:Guassian}, and Figure~\ref{fig:exp}. 


\begin{table*}[ht!]
\centering
\begin{tabular}{|c|c|c|c|c|c|c|c|c|c|c|}
\hline
Number of arms ($K$) & $50$  & $100$ & $150$ & $200$  & $250$ & $300$ & $350$ & $400$ & $450$ & $500$   \\ \hline
UCB (seconds) &  $2.13$  & $2.76$ & $3.52$ & $3.75$  & $4.29$ & $5.04$ & $6.44$ & $7.62$ & $8.74$ & $12.76$   \\ \hline
EG (seconds) &  $2.14$  & $2.51$ & $3.31$ & $4.35$  & $5.72$ & $6.36$ & $8.77$ & $9.77$ & $14.92$ & $19.12$ \\\hline
TS (seconds) &  $13.87$  &  $25.93$ & $38.49$ & $49.87$  &$60.78$ & $74.21$ & $87.84$ & $97.28$& $110.25$ & $120.38$  \\ \hline
KL-UCB (seconds) & $12.13$  & $20.11$ & $35.28$ & $40.31$ &  $49.91$  & $59.89$ & $65.17$ & $82.49$ & $92.21$& $101.57$    \\ \hline
MOSS (seconds) & $2.45$  &  $3.11$ & $4.27$ & $4.33$ & $5.36$  & $6.94$ & $7.47$ & $7.84$ & $9.00$ & $12.72$ \\ \hline
\end{tabular}
\caption{Average total computational time \protect\footnotemark}\label{table:TimeCompEx}
\vspace{-0.2cm}
\end{table*}
\footnotetext{All the experiments are implemented in
MATLAB on a desktop with an Intel Core i$7$-$11700$KF CPU and $16$GB of RAM. }


The performance evaluation reveals that KL-UCB and MOSS display similar levels of performance in terms of regret and cumulative reward. The arm selections of these two algorithms are also comparable, as illustrated in Figure~\ref{fig:exp_arms}. Notably, our proposed algorithms outperform traditional algorithms, such as EG and basic UCB, in both regret and cumulative reward. This outcome validates the theoretical results previously established.
While KL-UCB exhibits better regret performance compared to MOSS, as evidenced in Figures~\ref{fig:Uniform_regret} and \ref{fig:Guassian_regret}, MOSS proves to be more computationally efficient, as shown in Table~\ref{table:TimeCompEx}. This aligns with MOSS's design, which is intended to handle MAB problems with numerous arms and high dimensions. In contrast, KL-UCB incurs a higher computational cost due to the need to update the UCB for each arm at every time step. This process involves solving an optimization problem and computing the KL divergence between the estimated and prior distributions, making it more complex than MOSS's simpler sampling and updating methods.


Based on our simulation results, we have found that the TS algorithm performs optimally when dealing with uniform 
and Gaussian 
distributions. This is due to the simplicity and well-established mathematical representation of these distributions. Therefore, the TS algorithm is able to effectively model the uncertainty of the reward distribution and make informed decisions regarding the exploration-exploitation trade-off. However, when the buyer's valuation is drawn from an exponential distribution, as shown in Figure~\ref{fig:exp_avg_reward}, the performance of the TS algorithm is the poorest compared to the other algorithms tested. This can be attributed to the heavier tails of exponential distributions, which tend to concentrate the expected rewards in a small number of arms, making it more difficult for the TS algorithm to determine the best arm to play, as shown in Figure~\ref{fig:exp_arms}. In such scenarios, alternative algorithms such as EG, UCB, MOSS, or KL-UCB have been found to perform better as they adopt different strategies to handle the exploration-exploitation trade-off. 

\section{Conclusion and future work}
\label{summary}
In this paper, we presented two novel real-time online pricing mechanisms for allocating heterogeneous edge resources without requiring prior knowledge of demand distribution. 
To capture the preferences of delay-sensitive buyers, our proposed MAB model takes into consideration both the VM types and their geographic locations, while allowing multi-product purchases. 
The proposed algorithms with performance guarantees demonstrate efficient resource allocation and maximization of profit for the platform. 
The numerical results indicate the superiority of our proposed online mechanisms over the benchmark schemes derived from the traditional bandit literature.

\bibliographystyle{IEEEtran}

\bibliography{ref.bib}

\appendix

\subsection{Upper confidence bound (UCB) algorithm}
\label{UCB_alg}
The UCB algorithm \cite{UCB} provides a trade-off between exploration and exploitation by selecting the arm with the highest upper confidence bound. In the first $K$ rounds, the platform will play each price vector once for explorations. After round $K$, according to past observations till $t - 1$, the platform selects the price vector with the highest $\text{UCB}_{\bp,t-1}$ till that time. Here, the upper confidence bound (UCB) of price vector $p$ at time $t$ can be expressed as 
\begin{align} 
      \text{UCB}_{\bp,t} = \hat{r}^t(\bp)  + \sqrt{\frac{\log{t}}{n_{t}(\bp)}}
\end{align}
The UCB formula has two components: the empirical reward $\hat{r}^t(\bp)$ estimate up to time $t$, and a term that is inversely proportional to the number of times the arm has been played. This second term, $\sqrt{\frac{\log{t}}{n_{t}(\bp)}}$, helps to avoid always playing the same price vector without checking out others. As $n_{t}(\bp)$ increases, the UCB decreases, promoting exploration of other price vectors.

\begin{algorithm}[h!]
\caption{Upper confidence bound (UCB)}\label{Alg:UCB}
\begin{algorithmic}[1]
    \State Try each arm once
    \For{$t = K+1$ to $T$ (i.e., until resource exhausted)}
        \State pick a price vector $\bp^t = \argmax_{\bp \in \mathcal{P}} \text{UCB}_{\bp,t}$ 
        \State observe the consumption $\bc^t$ and reward $r^t$
    \EndFor
\end{algorithmic}
\end{algorithm}

\subsection{Thompson Sampling (TS) algorithm}
\label{TS_alg}
Thompson Sampling (TS) \cite{TS} is a popular Bayesian algorithm that uses a probabilistic approach to solve the MAB problem. It works by maintaining a posterior distribution over the expected reward of each arm and sampling an action from this distribution. This way, the algorithm gradually converges to the true expected reward of each arm, and the arm with the highest expected reward will be selected more often. This approach has been shown to be efficient in terms of regret, as it balances exploration and exploitation effectively. 

The algorithm starts by sampling from the prior distributions for each arm. At each time step, the arm with the highest sample is selected, and a reward is received. The posterior distribution for the selected arm is then updated using Bayesian updating, which involves incorporating the observed reward into the prior distribution to obtain a new estimate of the expected reward for that arm. The algorithm repeats this process until the end of the time horizon. Algorithm \ref{Alg:TS} provides the pseudocode of TS algorithm.

\begin{algorithm}[h!]
\caption{Thompson Sampling (TS)}\label{Alg:TS}
\begin{algorithmic}[1]
    \For{each arm $k = 1$ to $K$}
        \State {\!\!\!\!\!\!\!\!  Sample from the prior distribution for arm $k$}
    \EndFor
    \For{$t = 1$ to $T$ (i.e., until resource exhausted)}
        \State \!\!\!\!\!\!\!\! Choose arm $k$ with highest sample
        \State \!\!\!\!\!\!\!\! Set a price vector $\bp^t$ that corresponds to arm $k$
        \State \!\!\!\!\!\!\!\! Observe the consumption $\bc^t$ and reward $r^t$
        \State \!\!\!\!\!\!\!\! Update the posterior distribution using Bayesian updating
    \EndFor
\end{algorithmic}
\end{algorithm}

\subsection{Epsilon greedy (EG) algorithm}
\label{EG_alg}
The EG \cite{Epislongreedy} algorithm is a decision-making strategy that balances exploration and exploitation. The platform first generates a random number, and if this value, $\rho$, is less than the predefined $\epsilon$, a random action will be chosen. Otherwise, the platform will select the price vector with the highest empirical average reward. Algorithm \ref{Alg:Epsilon-greedy} provides the pseudocode of EG algorithm.

When $\epsilon$ is set to 0, the EG algorithm becomes a pure greedy algorithm without exploration. In this case, each price arm is selected with an equal probability. As $\epsilon$ increases, the frequency of random selection and exploration increases, leading to the optimal price vector being selected more often. At the same time, the non-optimal price vectors will be selected less frequently. When $\epsilon$ equals 1, the algorithm shifts towards a pure exploration strategy without exploitation. The price vectors are again selected randomly with equal probability, regardless of their estimated reward.

\begin{algorithm}[H]
\caption{Epsilon-greedy (EG)}\label{Alg:Epsilon-greedy}
\begin{algorithmic}[1]
    \State System Initialization: $\epsilon$
    \For{$t = 1$ to $T$ (i.e., until resource exhausted)}
    \State{\!\!\!Toss a coin with success probability $\epsilon$} 
    \State{\!\!\!\textbf{if} Success \textbf{then}}
        \State ~ explore: pick a price vector $\bp^t$ uniformly at random
    \State{\!\!\!\textbf{else}}
        \State ~ exploit: pick a price vector $\bp^t = \argmax_{\bp \in \mathcal{P}} \text{UCB}_{\bp,t}$ 
    \EndFor
\end{algorithmic}
\end{algorithm}

\end{document}